\pdfoutput=1

\documentclass[11pt]{article}

\usepackage[final]{acl}

\usepackage{times}
\usepackage{latexsym}

\usepackage[T1]{fontenc}

\usepackage[utf8]{inputenc}

\usepackage{microtype}

\usepackage{inconsolata}
\usepackage{multirow}
\usepackage{graphicx}
\usepackage{booktabs}
\usepackage{xspace}
\newcommand{\euandi}{\textsc{EUandI-2024}\xspace}

\newcommand{\mistral}{\textsc{Mistral}\xspace}
\newcommand{\mixtral}{\textsc{Mixtral}\xspace}
\newcommand{\llama}{\textsc{Llama-2-13B}\xspace}
\usepackage{enumitem}

\title{Investigating LLMs as Voting Assistants via Contextual Augmentation: \\
A Case Study on the European Parliament Elections 2024
}

\author{Ilias Chalkidis \\
        Department of Computer Science, University of Copenhagen, Denmark\\}

\begin{document}
\maketitle
\begin{abstract}
In light of the recent 2024 European Parliament elections, we are investigating if LLMs can be used as Voting Advice Applications (VAAs). We audit \mistral and \mixtral models and evaluate their accuracy in predicting the stance of political parties based on the latest ``EU and I'' voting assistance questionnaire. Furthermore, we explore alternatives to improve models' performance by augmenting the input context via Retrieval-Augmented Generation (RAG) relying on web search, and Self-Reflection using staged conversations that aim to re-collect relevant content from the model's internal memory. We find that \mixtral is highly accurate with an 82\% accuracy on average with a significant performance disparity across different political groups (50-95\%). Augmenting the input context with expert-curated information can lead to a significant boost of approx.~9\%, which remains an open challenge for automated RAG approaches, even considering curated content.
\end{abstract}

\section{Introduction}
\label{sec:intro}

While Large Language Models (LLMs) exhibit unprecedented Natural Language Understanding capabilities, recent work has been exploring political biases~\cite{feng-etal-2023-pretraining} and political reasoning capabilities~\cite{chalkidis-and-brandl-eu-llama-2024} in LLMs.\footnote{We present related work in Section~\ref{sec:related_work}.}  LLMs prove to have a considerable level of political knowledge and be capable of reasoning in this context. Hence, an open question is what is the potential of newer and bigger LLMs to disseminate political information and inform citizens, as an alternative, or in addition, to traditional media.

In pre-election periods, e.g., as in May 2024 upon the 2024 European Parliament elections in June, Voting Advice Applications (VAAs) have been used to inform the public about the political stance of candidate parties and help citizens find out which parties align the most with. VAAs rely on fixed (pre-defined) curated questionnaires, where citizens have to answer several multiple-choice questions, which are then parsed by a rule-based system to compute affinity to the candidate parties. While VAAs are good predictors relying on curated content by experts, they are static `one-size-fits-all' solutions covering a few topics with pre-defined questions that do not necessarily cover the interests of all individual users in detail. 

In this regard, conversational AI agents (LLMs) could potentially pose an interesting alternative for political information dissemination that better satisfies the interests of individual users with different priorities in a personalized contextualized responsive manner. Users could have open-ended conversations that help them cover more topics of interest in greater detail compared to traditional VAAs or other forms of media (e.g., news articles).\footnote{We present limitations, and ethical considerations (risks and open challenges) in Section~\ref{sec:limitations}.}

In this work, we ask the question: \emph{Could state-of-the-art LLMs potentially be used as VAAs?} And how, can we enhance their abilities with contextual augmentation? To answer this question, we employ two of the most recent powerful open-access LLMs, \mistral~\cite{jiang2023mistral} and \mixtral~\cite{jiang2024mixtral}, and audit their abilities as VAAs using a newly released dataset based on the ``EU and I 2024'' (\euandi) VAA questionnaire.\footnote{\url{https://euandi.eu/}}  Furthermore, we explore alternatives to improve models' performance by augmenting the input context via Retrieval-Augmented Generation (RAG) relying on web search, and ``Self-Reflection'' using staged conversations that aim to re-collect relevant content from the model's internal memory. 

\begin{figure*}[t]
    \centering
    \resizebox{\textwidth}{!}{
    \includegraphics{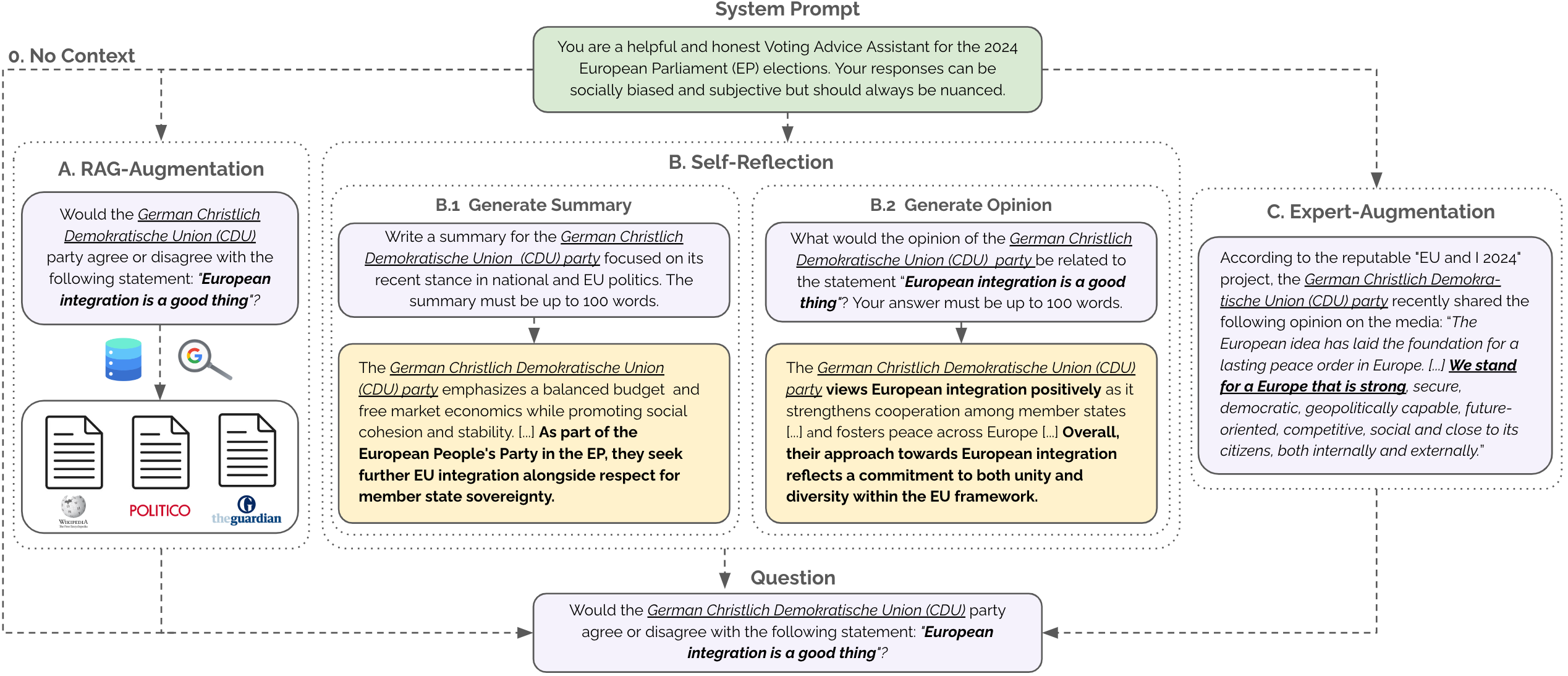}
    }
    \vspace{-6mm}
    \caption{Depiction of the experimental framework. In Setting (0), there is no context augmentation. In Setting (A) the context is augmented using web search to retrieve relevant content. In Setting (B), the context is \emph{self-augmented} by asking the model preliminary questions to generate a summary for the party and its expected opinion related to the question. In Setting (C), the input context is \emph{augmented} with the party's position related to the question.}
    \vspace{-5mm}
    \label{fig:framework}
\end{figure*}

We find that the larger LLM, \mixtral, is highly accurate with an 82\% accuracy on average while augmenting the input context with expert-curated ``gold-standard'' information can lead to a significant boost of approx.~9\%, which remains an open challenge for automated augmentation approaches, even considering curated content. Interestingly, RAG leads to a substantial performance boost in the case of \mistral (+8\%). Similar to \citet{chalkidis-and-brandl-eu-llama-2024}, we find a significant disparity in how accurately the models represent different political groups (50-95\%).\vspace{-1mm}

\section{Experimental Set Up}
\subsection{Dataset}
\label{sec:dataset}

We use the questionnaire from the ``EU and I 2024'' (\euandi) project as an evaluation benchmark. \euandi is a Voting Advice Application (VAA), released before the recent European Parliament elections to help EU citizens find their affinity to candidate parties. The questionnaire has 30 questions in the form of a political statement followed by 5 available options on Likert scale from complete disagreement to complete agreement. The \euandi project also provides the answers to the statements in question for all national parties across EU member states and euro-parties, alongside a verbatim justification, i.e., an excerpt from the party's program or public media statement identified by the project experts. 

As part of this work, we redistribute \euandi as a unified dataset, including the parties' answers, and justifications.\footnote{\url{https://huggingface.co/datasets/coastalcph/euandi_2024}} We include translated versions from the original language to English automatically derived using \mixtral (Appendix~\ref{sec:translate}).

\subsection{Examined Models}
\label{sec:models}
Across all experiments, we use two representative ``open-weight'' LLMs:  Mistral AI's 7B~\cite{jiang2023mistral}, dubbed \mistral, and 8$\!\times\!$7B Mixture of Experts (MoE)~\cite{jiang2024mixtral}, dubbed \mixtral, models. We use the latest chat-based versions for both models, i.e., Instruct-v0.2 (Dec 2023) and Instruct-v0.1 (Dec 2023), respectively.

\begin{figure*}[t]
    \centering
    \resizebox{\textwidth}{!}{
    \includegraphics{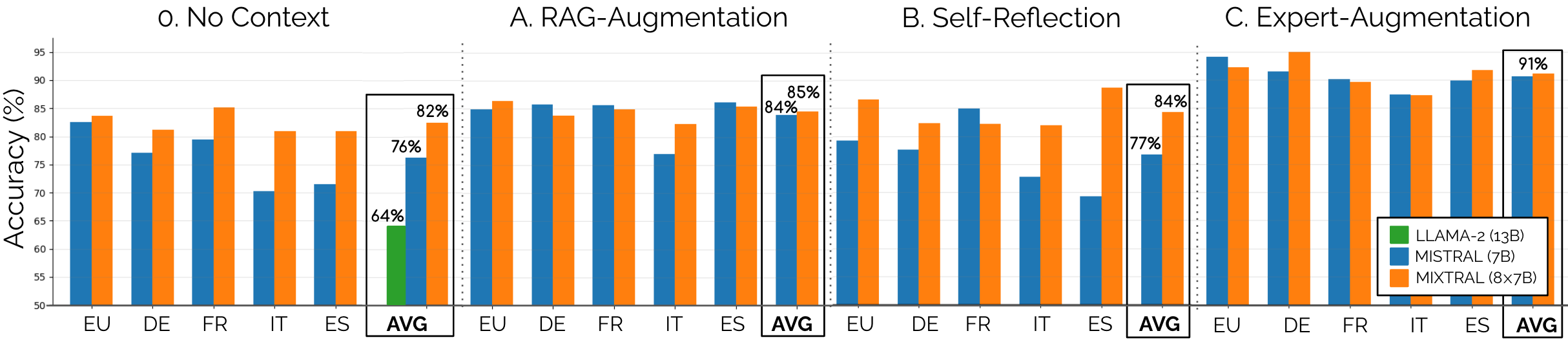}
    }
    \vspace{-6mm}
    \caption{Accuracy of the examined models (\mistral in blue, and \mixtral in orange) on \euandi dataset across all settings (Section~\ref{sec:settings}) and examined groups (4 EU Member States + euro-parties).}
    \label{fig:main_results}
    \vspace{-4mm}
\end{figure*}

\subsection{Auditing Settings} 
\label{sec:settings}
We audit the models as Voting Assistant Applications (VAAs) in 4 different settings (Fig.~\ref{fig:framework}):

\paragraph{0. No Context:} In this case, we ask the model to speculate how a specific political party, e.g., the German Christlich Demokratische Union  (CDU) party, would reply to a given question from the \euandi questionnaire with no additional context, to assess its capabilities to reason solely relying on its internal knowledge. The only context available is the system prompt (green box in Fig.~\ref{fig:framework}) and the question (last purple box).

\paragraph{A. RAG-Augmentation:} In this case (part A of Fig.~\ref{fig:framework}), we use Retrieval-Augmented Generation (RAG)~\cite{lewis-rag-2020} with the retrieval engine of \emph{HuggingChat}\footnote{\url{https://huggingface.co/chat/}} which relies on web search. First, the system retrieves relevant indexed documents (pages) from the web using the question as a query. In this case, the model can rely on both its internal knowledge and external knowledge retrieved via the RAG system to answer the question.

\paragraph{B. Self-Reflection:} In this case (part B of Fig.~\ref{fig:framework}), we perform a 3-step staged conversation (dialog) with the model. First, we ask the model to provide a summary of the relevant political party. Secondly, we ask the model to speculate the party's stance related to the political statement in question. In both steps, the model solely relies on its internal knowledge, i.e., no retrieval engine enhances the context with external knowledge. Lastly, we ask the model to answer the question in light of the 2 preceding question-answers pairs. This can be seen as a form of guided Chain-of-Thought (CoT)~\cite{Wei2022ChainOT}, where the user asks relevant preliminary questions to prepare (contextualize) the model for answering the related final question. 

\paragraph{C. Expert-Augmentation:} In this case (part C of Fig.~\ref{fig:framework}), we use the expert-curated ``gold-standard'' justification of the relevant party from the \euandi dataset (Section~\ref{sec:dataset}) related to the follow-up question to augment the model's context. This setting can be seen as Retrieval-Augmented Generation (RAG)~\cite{lewis-rag-2020} utilizing an \emph{Oracle} retriever that always retrieves the ``perfect'' relevant text snippet (source) related to the question in hand.

\subsection{Use of Data} We audit the models in English using the \euandi dataset (Section~\ref{sec:dataset}) considering the answers and justifications (opinions) of the top 5 most popular parties from Germany, France, Italy, and Spain based on the latest national opinion polling (15/05/2024). These are the 4 most populous EU member states with the most Members of the European Parliament (MEPs). 
We also consider all euro- (EU-level) parties that are represented in the European Parliament. These are 27 parties in total.\footnote{We present additional details in Appendix~\ref{sec:data_details}.}

\subsection{Evaluation Protocol}

We measure models' performance using accuracy, i.e., the percentage of questions where the model generates the same answer as the examined party. Following \citet{chalkidis-and-brandl-eu-llama-2024}, we binarize parties' answers and model responses into two groups: agreement and disagreement, since the different shades of both groups, e.g., ``tend to agree'' and ``strongly agree'', are susceptible to sensitive interpretations, i.e., the intensity of agreement/disagreement instead of the general stance.


\section{Results}

\subsection{Main Results}

In Fig.~\ref{fig:main_results}, we present the main results for the four alternative settings 0, A-C (Section~\ref{sec:settings}). First, we observe that across settings \mixtral outperforms \mistral, especially when solely relying on its internal knowledge (Set.~0: 82\% vs 76\%), which is in line with the findings in general NLU benchmarking~\cite{jiang2024mixtral}. This is highly expected since \mixtral is a much bigger model with 8$\!\times$ more total trained parameters and 2$\!\times$ active parameters used on inference time.

We find that \textsc{Llama-2 (13B)} of \citet{touvron2023llama} -released July 2023- used in the study of \citet{chalkidis-and-brandl-eu-llama-2024}  has an accuracy of 64\%; six months later \mistral and \mixtral score +12\% and +18\%, respectively. Concerning performance disparity across groups in Setting 0, we observe that the performance for French (FR) and euro- (EU) parties is approx.~85\% for \mixtral, while for German (DE), Italian (IT) and Spanish (ES) parties is approx~81\%. 

Considering the results in Set.~A (Fig.~\ref{fig:main_results}), we observe that augmenting the context with a web-based RAG leads to minor performance improvements (approx.~3\%) for \mixtral. The performance improves approx.~5\% for the worst performing EU member state Spain (ES). Contrary, the smaller \mistral, has a significant performance boost, approx.~8\% on average. The most commonly used web sources across all parties were: Wikipedia (14\%), Politico (6\%), and ``The Guardian'' (5\%).

Considering the results in Set.~B (Fig.~\ref{fig:main_results}), self-augmenting the context leads to minor performance improvements compared to Set.~0, except for Spain (ES) with a boost of approx.~+8\%. In this case (Set.~B), similar to Set.~0, \mistral is limited to its own knowledge capabilities and is out-performed by 7\% from \mixtral (77\% vs 84\%).

\begin{figure}[h]
    \centering
    \resizebox{\columnwidth}{!}{
    \includegraphics{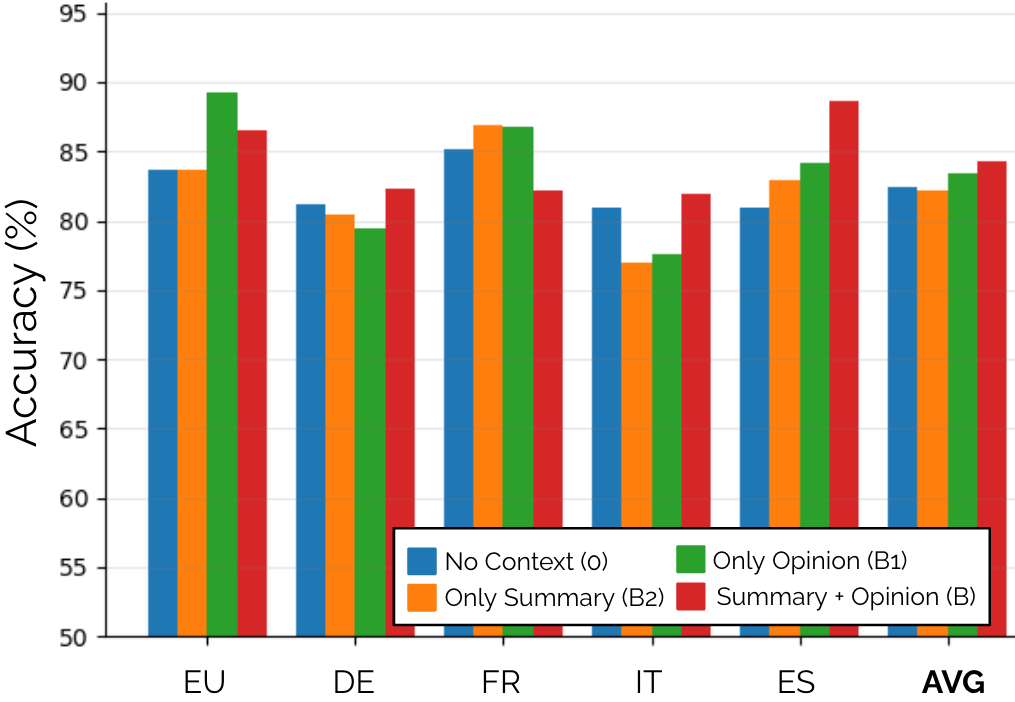}
    }
    \vspace{-7mm}
    \caption{Accuracy of \mixtral on different sub-settings of Setting B: Self-Augmented Context.}
    \label{fig:ablation}
    \vspace{-4mm}
\end{figure}

In Fig.~\ref{fig:ablation}, we present results of \mixtral, our best-performing model, by ablating the context used in Set.~B. We observe that inferring the party's opinion (Set.~B2: green bars) relevant to the question is more beneficial than inferring a general party summary (Set.~B1: orange bars), but combining both (Set.~B) leads to the best overall results.\vspace{-2mm}

Considering the results in Set.~C (Fig.~\ref{fig:main_results}), where the context is augmented with expert-curated ``gold-standard'' information, we observe a substantial performance improvement of approx.~15\% for \mistral (91\%), and 9\% for \mixtral (91\%) compared to Set.~0. This finding signifies that when ``gold-standard'' information is utilized as context, performance is improved substantially.

Since automated augmentation (Set.~A-B) leads to minor improvements for \mixtral, contrary to expert-curated content (Set.~C), we consider the use of curated domain-specific corpora for RAG, in contrast to open-ended (web-based). 

\subsection{Improving RAG with Curated Corpus}
\label{sec:custom_rag}

\begin{figure}[t]
    \centering
    \resizebox{\columnwidth}{!}{
    \includegraphics{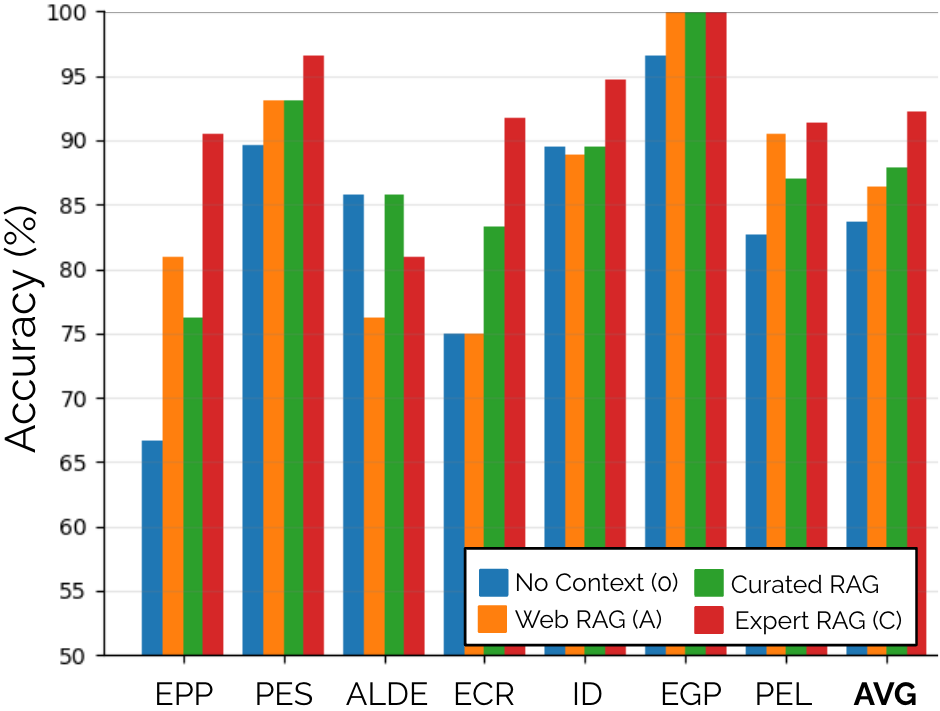}
    }
    \vspace{-7mm}
    \caption{Accuracy of \mixtral using RAG based on different corpora (document collections).}
    \label{fig:rag_ablation}
    \vspace{-6mm}
\end{figure}

To test this hypothesis, we create a small curated corpus for retrieval, where the only available sources are paragraphs from the official euro-parties' pre-election \emph{manifestos}.\footnote{We present additional details on Appendix~\ref{sec:custom_rag_more}.}  In Fig.~\ref{fig:rag_ablation}, we present the results for the 7 euro-parties. We observe that results vary a lot across euro-parties, but there is a minor performance improvement of 1.5\% using the curated (green bars) vs web-based (orange bars) RAG. Performance improvements with the curated RAG are more vibrant for the worst-performing parties (EPP, ECR). By manually inspecting the retrieved paragraphs, we observe that there is not always highly relevant content, which limits the potential of this approach.

\begin{figure}[h]    \centering
    \resizebox{\columnwidth}{!}{
    \includegraphics{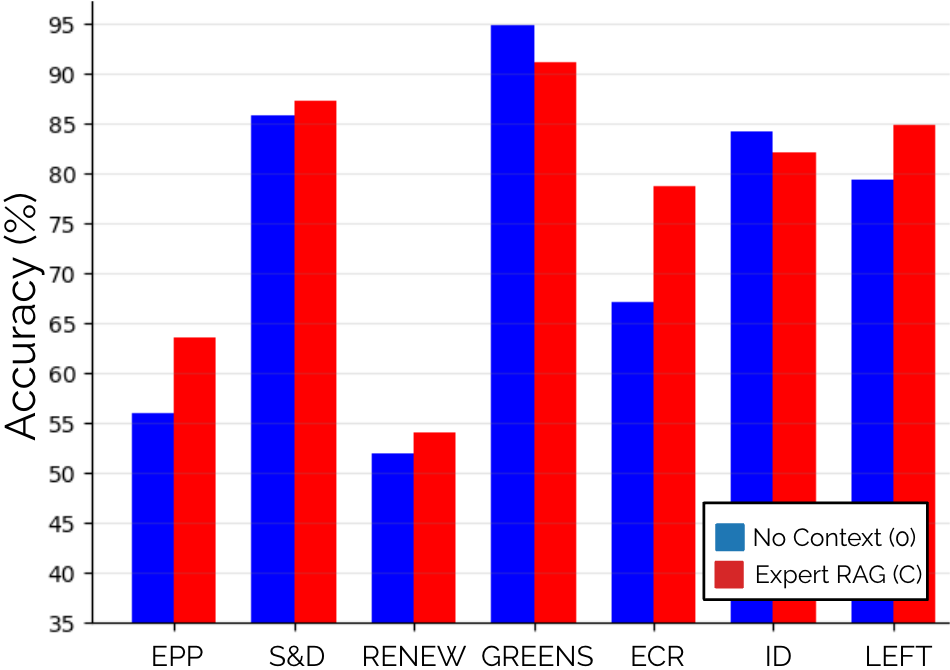}
    }
    \vspace{-6mm}
    \caption{Accuracy of \mixtral across euro-groups, based on the coalitions formed in the 9th European Parliament (2019-2024).}
    \label{fig:euro_groups}
    \vspace{-5mm}
\end{figure}

\subsection{Breakdown by Euro-group}

In Fig.~\ref{fig:euro_groups}, we present the results for \mixtral now grouped by euro-group alliances, i.e., European Parliament groups that represent different political ideologies and agendas. Similar to~\citet{chalkidis-and-brandl-eu-llama-2024}, we observe that there is a large performance disparity, with higher accuracy when it comes to groups with distinct political leanings, e.g., the Greens or the right-wing ID, compared to others, e.g., centre-right parties (EPP, RENEW), and the euro-skeptic ECR. Expert-augmented context leads to substantial performance improvements for parties affiliated with EPP (+5\%) and ECR (+15\%), which align with the results in Section~\ref{sec:custom_rag}.

\subsection{Qualitative Results}
In Tables~\ref{tab:examples_0}-\ref{tab:examples_c} in Appendix~\ref{sec:examples}, we present responses by the best-performing model, \mixtral, across settings, considering examples of both correct and mis- predictions. We observe that the model answers the question (pick an option) and then provides a brief justification to support its decision. As demonstrated in Tables~\ref{tab:examples_a} and \ref{tab:examples_c}, in certain cases, where the context is augmented with additional content, it correctly adjusts its answer, comparing to no content (Table~\ref{tab:examples_0}), while in other cases not.

\section{Related Work}
\label{sec:related_work}

\citet{feng-etal-2023-pretraining} were the first to explore the political biases of prior-art (smaller, not instruction-tuned, human-feedback aligned) language models, such as \textsc{BERT} and \textsc{GPT-2}, showcasing that models have different political leanings based on the political compass covering a 2-dimensional political spectrum (left/right, authoritarian/libertarian) heavily focused on the US political context. Furthermore, they demonstrated how these biases affect downstream performance in critical tasks, e.g., toxicity classification and fake news detection, and how continuous pre-training on politically biased data can further exacerbate these biases. 

\citet{Santurkar2023} examined the questions of whose opinions are reflected by LLMs using the  Pew Research Center's ``American Trends Panel'' (ATP) questionnaire and survey. They found that base (pretrained-only) LLMs, like \textsc{GPT-3}, mostly align with lower-income, moderate, catholic groups. Contrary, recent human-feedback-aligned instruction-tuned LLMs, like \textsc{GPT-3.5}, align more with high-income, liberal, well-educated, not religious groups, and thus are in general less representative of overall opinions.

\citet{chalkidis-and-brandl-eu-llama-2024} are the first to explore the political biases of LLMs in the context of EU politics. They audited the chat-based \llama on the \textsc{EUandI-2019} dataset \cite{euandi} showcasing that the model has limited knowledge of parties' stances with significant disparity across political groups. They also adapted the model to partisan data demonstrating a high degree of re-alignment in most cases. 

\citet{Hartmann2023ThePI} conducted a similar analysis focusing on ChatGPT. They audited the model based on German and Dutch national questionnaires, overall coming to a similar conclusion as \citet{feng-etal-2023-pretraining} and \citet{chalkidis-and-brandl-eu-llama-2024} that ChatGPT leans mostly left-libertarian. In our work, we focus on and investigate the potential of LLMs to disseminate political information and be used as VAAs accurately, not to examine the models' political leanings out-of-context.

\citet{rottger-etal-2024-political} scrutinize the shortcomings of the unrealistic closed-form ``forced'' evaluation, e.g., multi-choice questionnaires, used in the related work on evaluating values and opinions in LLMs, such as political biases. They demonstrated the lack of robustness, and compliance in models' answers when questioned in open-ended settings. They advocate for application-level evaluations, the use of robustness tests, and the framing of claims concerning only the examined models.

Given the aforementioned background, our work explores the under-studied EU political context. For this, we rely on up-to-date (May 2024) political content (information), namely the \euandi questionnaire, and recently developed ``open-weight'' LLMs that give us the flexibility to examine different auditing settings. To the best of our knowledge, we are the first to investigate the use of contextual augmentation via RAG in the context of political NLP studies~\cite{politicalnlp-2024-natural}.

\section{Conclusion and Future Work}

In conclusion, we find that recent larger LLMs have a great potential to serve as Voting Advice Applications (VAAs) with an accuracy of 80\%+ out-of-the-box. Nonetheless, there is a substantial performance disparity among different political groups. The overall performance is improved when we consider contextual augmentation with expert-curated information (90\%+). Unfortunately, we find that automatically augmenting the context via web-based RAG is still an open challenge leading to minor improvements (+3\%) for \mixtral; contrary web-based RAG substantially boosts the performance of the smaller \mistral (+8\%). 

In future work, we will further investigate the use of custom-curated RAG, which has the potential to further improve performance by grounding models' generation on useful up-to-date credible sources. We will also investigate the potential benefit of continuous instruction-tuning of LLMs with partisan data, such as the ``EU Debates'' corpus of \citet{chalkidis-and-brandl-eu-llama-2024}. Similarly, a usability study assessing the experience of human actors when interacting with LLMs in the context of politics is a crucial next step.

\section*{Limitations}
\label{sec:limitations}

\paragraph{EU Scope} The scope of our work is limited in the recent EU political context (national parties from 4 EU member states, and the European Parliament Elections 2024) in English language, so all findings and claims are subjected to this specific scope. 

\paragraph{Qualitative analysis} Our study is also limited to mostly quantitative results, e.g., how many times the models replied with the right option, i.e., multi-class decision. Analyzing the models' responses to investigate their argumentation for justifying (explaining) the given answers is also an important aspect related to trustworthiness and robustness~\cite{rottger-etal-2024-political}. From a usability perspective, LLMs knowledge should be better understood as \textit{justified} true beliefs, i.e., ``$x$ thinks that $p$, $p$ is true, and $x$ has some justification for thinking it''~\cite{Nozick2000-NOZKA}, instead of any other definition that does not directly involve the ability to justify; since the value of disseminating information (knowledge) is limited without reasonable justification (explanation). \citet{fierro-etal-2024-knowledge} present a thorough survey on interpretations of knowledge definitions applicable to LLMs.

\paragraph{Usability} As our work investigates the potential of LLMs as voting assistants, a usability study where we assess users' experience while interacting with the examined LLMs would provide valuable insights. In the lack of resources, we omit this aspect for a future study. 

\paragraph{Models} In our study, we considered 2 representative ``open-weight'' chat-based LLMs: \mistral 7B, and \mixtral 8$\!\times$\!7 using a single NVIDIA A100 (40G) GPU card. \mixtral is the 3rd best-performing ``open-weight'' model according to the up-to-date (15 June 2024) results in the HELM Leaderboard~\cite{liang2023holistic}.\footnote{\url{https://crfm.stanford.edu/helm/mmlu/latest/}} We lack the resources to investigate the 2 best, much bigger, models: Meta's Llama-3 70B (released Apr 2024), and the recent Mixtral 8$\!\times\!$22B (released May 2024).

\paragraph{Indexed web-based RAG} Most web-based RAG systems rely on an indexed (older) version of the web rather than an online version retrieving the most recent pages (documents). This can be an issue in time-sensitive applications like the one we explore, since the most up-to-date information, e.g., parties' manifestos, recent interviews, and talks, may not be part of the library.

\section*{Ethical Considerations}
\label{sec:ethics}

While we firmly believe that conversational AI agents (LLMs) could potentially pose an interesting alternative for political information dissemination, several notable risks and challenges should be taken into consideration and mitigated before such critical applications are considered for deployment in public and popularized:
\begin{itemize}[itemsep=-0.2em,leftmargin=8pt]
    \item  First and foremost, LLMs have intrinsic political biases~\cite{feng-etal-2023-pretraining, chalkidis-and-brandl-eu-llama-2024}, which was also a finding in our work, which means that they better represent specific political groups and ideologies over others. Political \emph{misrepresentation} can be seen as a form of \emph{misinformation}~\cite{chen2023combating}, which is a critical issue. It should be also understood as a form of lack of \emph{fairness} since some political groups are better (more accurately) served (represented) than others. Nonetheless, misinformation and misrepresentation, which undermine democracy, are great open challenges for human societies -and traditional media- in general.
    \item Secondly, political information dissemination as a sensitive application with critical implications in human societies must be grounded in reliable sources and such sources should accompany the model's generated text so that users can read them and fact-check information. Retrieval-Augmented Generation (RAG) can be a potential direction to tackle this challenge, but in our work, we find that automated RAG-based solutions do not boost performance. Given the promising results we have with the expert-curated augmentation, we believe that the use of curated domain-specific corpora for RAG can be a potential solution, limiting retrieval to highly relevant and credible sources. Moreover, explainability in the era of LLMs should be investigated to help users identify how the retrieved sources, alongside the original training data themselves, led to the generated content.
\end{itemize}

In general, the use of conversational AI agents (LLMs) as Voting Advice Applications (VAAs) is a challenging sensitive topic and should be further investigated. In this direction, collaboration with political scientists, and representatives from the political parties is of paramount importance to secure an operational framework under which LLMs will be considered a credible curated source for political information dissemination.

\section*{Acknowledgements}
We thank our colleagues at the CoAStaL NLP group Stephanie Brandl, Constanza Fierro, and Danae Sánchez for their valuable comments. This project is is funded by the Novo Nordisk Foundation (grant NNF 20SA0066568).

\bibliography{custom}

\appendix

\begin{table*}[t]
    \centering
    \resizebox{\textwidth}{!}{
    \begin{tabular}{c|c|c|c|c|c|c|c|c|c|c|c|c|c|c|c|c|c|c|c}
        \toprule
       \multicolumn{5}{c|}{EU} & \multicolumn{5}{c|}{DE} & \multicolumn{5}{c|}{IT} & \multicolumn{5}{c}{ES}\\
       \midrule
               Party & 0 & A & B & C & Party & 0  & A & B & C & Party & 0  & A & B & C & Party & 0 & A & B & C \\ 
        \midrule
       EPP & 66.7 & 81.0 & 81.0 & 90.5 & CDU & 69.0 & 75.0 & 69.0 & 100.0& FI & 64.0 & 68.0 & 76.0 & 84.0 & PP & 80.8 & 88.5 & 92.3 & 84.6\\
       \midrule
       ECR & 75.0 & 75.0 & 83.3 & 91.7 & \multicolumn{5}{c|}{-}  & FDI & 59.3 & 68.0 & 74.1 & 77.8 & \multicolumn{5}{c}{-} \\
       \midrule
       PES & 89.7 & 93.1 & 89.7 & 96.6 & SPD & 82.8 & 86.2 & 86.2 & 93.1 & PD & 85.7 & 85.7 & 85.7 & 85.7 & PSOE & 81.5 & 88.5 & 85.2 & 92.6 \\
      \midrule
      ALDE & 85.7 & 76.2 & 85.7 & 81.0 & FDP & 70.0 & 80.0 & 76.7 & 83.3 & M5S & 93.1 & 69.0 & 86.2 & 89.7 & \multicolumn{5}{c}{-} \\
       \midrule
       EGP & 96.6 & 100.0 & 89.7 & 100.0& Grüne & 96.4 & 96.3 & 92.9 & 100.0 & AVS & 96.6 & 93.1 & 93.1 & 100.0 & AR & 84.6 & 91.7 & 92.3 & 96.2\\
        \midrule
        ID & 89.5 & 10.5 & 89.5 & 94.7& AfD & 86.2 & 89.7 & 86.2 & 93.1 & Lega & 86.7 & 83.3 & 76.7 & 86.7 & Vox & 86.2 & 75.9 & 86.2 & 96.6\\
       \midrule
       PEL & 82.6 & 90.5 & 87.0 & 91.3 & Linke & 82.8 & 75.9 & 82.8 & 100.0 & \multicolumn{5}{c|}{-}  & Sumar & 75.0 & 87.5 & 83.3 & 87.5 \\
         \midrule
       Avg. & 83.7 & 86.4 & 86.5 & 92.2 & Avg. & 81.2 & 83.8 & 82.3 & 94.9 & Avg. & 80.9 & 82.3 & 82.0 & 87.3 & Avg. & 81.2 & 85.4 & 88.7 & 91.2 \\
       \bottomrule
    \end{tabular}
    }
    \vspace{-2mm}
    \caption{Accuracy of \mixtral on \euandi dataset across all settings for all parties from different EU Member States. National parties are in the same row with the euro-party (first column) they affiliate with.}
    \label{tab:my_label}
\end{table*}

\begin{table*}[t]
    \centering
    \resizebox{\textwidth}{!}{
    \begin{tabular}{c|p{17cm}}
    \toprule
    Setting & Dialogue Template \\
    \midrule
    \textbf{0} & \textbf{User:} Would the \emph{[ORIGIN]} \emph{[PARTY NAME]} party agree or disagree with the statement "\emph{[STATEMENT]}"? \\
    & \textbf{Model:} The \emph{[ORIGIN]} \emph{[PARTY NAME]} party aligns with option ([...] \\
    \midrule
    \textbf{A} & \textbf{User:} Would the \emph{[ORIGIN]} \emph{[PARTY NAME]} party agree or disagree with the statement "\emph{[STATEMENT]}"? \\
    & \textbf{User:} [RETRIEVED WEB SNIPPETS]\\
    & \textbf{Model:} The \emph{[ORIGIN]} \emph{[PARTY NAME]} party aligns with option ([...] \\
    \midrule
    \textbf{B} & \textbf{User:} Write a summary for the \emph{[ORIGIN]}  \emph{[PARTY NAME]} party focused on its recent stance in national and EU politics. The summary must be up to 100 words. \\
    & \textbf{Model:} The \emph{[ORIGIN]} \emph{[PARTY NAME]} party [...] \\
    &  \textbf{User:} What would be the position of the \emph{[ORIGIN]} \emph{[PARTY NAME]} party related to the statement "\emph{[STATEMENT]}"? Your answer must be up to 100 words. \\
    & \textbf{Model:} The \emph{[ORIGIN]} \emph{[PARTY NAME]} party [...] \\
    & \textbf{User:} Would the \emph{[ORIGIN]} \emph{[PARTY NAME]} party agree or disagree with the statement "\emph{[STATEMENT]}"? \\
    & \textbf{Model:} The \emph{[ORIGIN]} \emph{[PARTY NAME]} party aligns with option ([...] \\
    \midrule
    \textbf{C} & \textbf{User:} According to the "EU and I 2024" project, the \emph{[ORIGIN]} \emph{[PARTY NAME]} party recently shared the following opinion on the media: "\emph{[JUSTIFICATION]}". \\
     & \textbf{User:} Would the \emph{[ORIGIN]} \emph{[PARTY NAME]} party agree or disagree with the statement "\emph{[STATEMENT]}"? \\
    & \textbf{Model:} The \emph{[ORIGIN]} \emph{[PARTY NAME]} party aligns with option ([...] \\
    \bottomrule
    \end{tabular}
    }
    \vspace{-2mm}
    \caption{Dialogue templates used across the alternative auditing settings (Section~\ref{sec:settings}).}
    \label{tab:templates}
    \vspace{-2mm}
\end{table*}

\section{Detailed Results}

In Table~\ref{tab:my_label}, we present results for all examined national parties (Section~\ref{sec:data_details}) across all examined settings (Appendix~\ref{sec:settings}). In Table~\ref{tab:ablation}, we present results for the ablation study of Setting C. In Table~\ref{tab:ablation_rag}, we present results for the ablation study considering different RAG systems.

\begin{table}[t]
\centering
\resizebox{\columnwidth}{!}{
\begin{tabular}{l|c|c|c|c|c|c}
\toprule
\bf Context Used & EU & DE & FR & IT & ES & Avg. \\
\midrule
None & 83.7 & 81.2 & 85.2 & 80.9 & 81.0 & 82.2 \\
Summary (S) & 83.6 & 80.5 & 86.8 & 76.9 & 82.9 & 81.9 \\
Opinion (O) & 89.2 & 79.4 & 86.8 & 77.5 & 84.1 & 83.4 \\
Both (S+O) & 86.5 & 82.3 & 82.2 & 82.0 & 88.7 & 84.5 \\
\bottomrule
\end{tabular}
}
\vspace{-2mm}
\caption{Accuracy of \mixtral on different sub-settings of Setting B: Self-Augmented Context.}
\vspace{-4mm}
\label{tab:ablation}
\end{table}

\begin{table}[t]
\centering
\resizebox{\columnwidth}{!}{
\begin{tabular}{l|c|c|c|c|c|c|c|c}
\toprule
\bf RAG  & EPP & PES & ALDE & ECR & ID & EGP & PEL & Avg. \\
\midrule
None & 66.7 & 89.7 & 85.7 & 75.0 & 89.5 & 96.6 & 82.6 & 83.7 \\
Web & 81.0 & 93.1 & 76.2 & 75.0 & 88.9 & 100.0 & 90.5 & 86.4 \\
Curated & 76.2 & 93.1 & 85.7 & 83.3 & 89.5 & 100.0 & 87.0 & 87.8 \\
Oracle  & 90.5 & 96.6 & 81.0 & 91.7 & 94.7 & 100.0 & 91.3 & 92.2 \\
\bottomrule
\end{tabular}
}
\vspace{-2mm}
    \caption{Accuracy of \mixtral using RAG based on different corpora (document collections).}
\label{tab:ablation_rag}
\vspace{-6mm}
\end{table}

\section{Parties examined in our study}
\label{sec:data_details}
We audit the models for the top 5 most popular -based on the latest national opinion polling (15/05/2024)- parties from Germany, France, Italy, and Spain. These are the following: 
\begin{itemize}[itemsep=-0.2em,leftmargin=8pt]
    \item German national parties: Christlich Demokratische Union (CDU), Sozialdemokratische Partei Deutschlands (SPD), Alternative für Deutschland (AfD), Freie Demokratische Partei (FDP), Die Linke, and Die Grünen.
    \item French national parties: Renaissance (RE), Rassemblement National (RN), La France Insoumise (LFI), Les Écologistes – Europe Écologie Les Verts (EELV), Parti Socialiste (PS), and Les Républicains (LR).
    \item Italian national parties: Lega Salvini Premier (Lega), Partito Democratico (PD), Fratelli D'Italia (FDI), Movimento 5 Stelle (M5S), Forza Italia (FI), and Alleanza Verdi e Sinistra (AVS).
    \item Spanish national parties: Partido Socialista Obrero Español (PSOE), Partido Popular (PP), Vox, Ahora Repúblicas, Sumar, and Podemos.
\end{itemize}

We also consider all euro-parties that were represented in the 9th European Parliament (2019-2024): European People's Party (EPP), Party of European Socialists (PES), Alliance of Liberals and Democrats for Europe Party (ALDE), European Conservatives and Reformists Party (ECR), European Green Party (EGP), Identity and Democracy (ID), and the Party of the European Left (PEL). We conduct all experiments in English. 

\section{Technical Details} 

\subsection{Machine-Translation}
\label{sec:translate}

Since all our experiments are in English, we translated the \euandi (Section~\ref{sec:dataset}) parties' justification from the original language to English automatically derived using \mixtral with the following instruction: \emph{``You are a helpful professional translation assistant. The translations must be faithful and accurate at a professional level meeting the highest European Commission official standards. Translate the following text from [SOURCE LANGUAGE] to English: [JUSTIFICATION]''}. The translated versions are part of the released dataset.\footnote{\url{https://huggingface.co/datasets/coastalcph/euandi_2024}}

\subsection{Curated RAG Implementation}
\label{sec:custom_rag_more}

We create a custom-curated corpus for document retrieval, where the only available sources are the official pre-election \emph{manifestos} (programs) of the 7 examined euro-parties. We automatically split all manifestos into paragraphs, which leads to 60-100 paragraphs per euro-party depending on the length of their manifesto. Then, we embed the paragraphs with the use of one of the best open-source document embedders, \texttt{sentence-transformers/all-mpnet-base-v2}, which is based on MPNet~\cite{mpnet2020} and fine-tuned in many sentence retrieval datasets, released by \citet{reimers-2019-sentence-bert}. We then retrieve the top 3 relevant paragraphs from each euro-party's manifesto using as query the statement in question, e.g., \emph{``The European Union should be enlarged to include Ukraine''}, and augment the context using the following template \emph{``These are relevant snippets from the official manifesto of the \underline{[ORIGIN]} \underline{[PARTY NAME]} party for the European Elections 2024: 1. \underline{[PARAGRAPH 1]} 2. \underline{[PARAGRAPH 2]} 3. \underline{[PARAGRAPH 3]}''}. 

\subsection{Dialogue Templates}

In Table~\ref{tab:templates}, we present the dialogue templates across settings (Section~\ref{sec:settings}). We prepend a short part of the model's expected response to enforce compliance with our instructions (questions). Without ``premeditating'' the model's response, we risk the lack of compliance due to (a) the model's learned guardrails from AI alignment, e.g., the model will not respond to questions related to sensitive topics or avoid sharing opinions, (b) invalid options, e.g., the model will return an invalid option (not a-e). Also, we eliminate the complexity of post-processing free-text responses, i.e., the response may lack references to the alphabetic option (a-e).

\section{Examples}
\label{sec:examples}

In Tables~\ref{tab:examples_0}-\ref{tab:examples_c}, we present responses by the best-performing model, \mixtral, considering examples of both correct and mis- predictions.

\begin{table*}[t]
    \centering
    \resizebox{\textwidth}{!}{
    \begin{tabular}{p{15cm}}
    \toprule
    \footnotesize{\textbf{Question}: Would the Christlich Demokratische Union Deutschlands (CDU) party agree or disagree with the statement: "Gender quotas (e.g. a minimum share of positions to be filled by women) are to be welcomed"?} \\
\footnotesize{\textbf{Model Answer}: The Christlich Demokratische Union Deutschlands (CDU) aligns with \underline{option (b)} - tends to disagree with the statement. The CDU believes in equal opportunities and equal representation, but they generally prefer voluntary measures to increase the number of women in leadership positions over gender quotas. However, it's essential to understand that individual attitudes within the party can vary, and the CDU's stance may evolve over time due to the ongoing discussion on gender quotas in politics and corporate settings both in Germany and the European Union.} \\
\footnotesize{\textbf{Party Answer}: (d) Tend to agree} \\
    \midrule
    \footnotesize{\textbf{Question}: Would the German Alternative für Deutschland (AfD) agree or disagree with the statement: "Immigration into European Union should be made more restrictive"?} \\
    \footnotesize{\textbf{Model Answer}: The German Alternative für Deutschland (AfD) party aligns with \underline{option (e)} - the party completely agrees with the statement "Immigration into European Union should be made more restrictive". The AfD has consistently advocated for stricter immigration policies and has been critical of the EU's handling of the migration crisis. Nevertheless, I recommend visiting AfD's official resources for the most up-to-date and accurate} \\
    \footnotesize{\textbf{Party Answer}:  (e) Completely agree} \\
    \midrule
    \footnotesize{\textbf{Question}: Would the German Die Linke (Linke) agree or disagree with the statement: "European integration is a good thing"?} \\
    \footnotesize{\textbf{Model Answer}: The German Die Linke (Linke) party aligns with \underline{option (b)}. They tend to disagree with the statement that European integration is a good thing. The party is critical of some aspects of European integration and believes that the current form of the European Union (EU) has failed to address the social and economic inequalities between its member states. They advocate for a more democratic, socially equitable, and peaceful Europe but with greater respect for national sovereignty. Nevertheless, the party does not reject the idea of European integration entirely, and their stance may evolve depending on the context and issues at hand.} \\
    \footnotesize{\textbf{Party Answer}: (e) Completely agree} \\
    \midrule
    \footnotesize{\textbf{Question}: Would the French Renaissance (RE) agree or disagree with the statement: "Taxation on the wealthiest part of the population should be increased"?} \\
    \footnotesize{\textbf{Model Answer}: The French Renaissance (RE) party aligns with \underline{option (a)} - they completely disagree with the statement "Taxation on the wealthiest part of the population should be increased." The party, led by President Emmanuel Macron, advocates for reducing taxes, particularly on businesses and high-earning individuals, to attract investment and stimulate economic growth. However, they also believe in implementing targeted welfare programs to help lower-income households and encourage social mobility. It is essential to note that perspectives might change depending on specific contexts and evolving economic conditions.} \\
    \footnotesize{\textbf{Party Answer}: (a) Completely disagree} \\
\midrule
\footnotesize{\textbf{Question}: Would the Identity and Democracy (ID) party agree or disagree with the statement: "The European Union should strengthen its security and defence policy"?} \\
\footnotesize{\textbf{Model Answer}: The Identity and Democracy (ID) party aligns with \underline{option (a)} - they completely disagree with the statement "The European Union should strengthen its security and defence policy." The ID party is eurosceptic and often critical of further integration in the EU. When it comes to security and defence, the party emphasizes national sovereignty and prefers intergovernmental cooperation over supranational policies. Therefore, they would likely argue that member states alone or through NATO should handle European defence policy, not the European Union.} \\
\footnotesize{\textbf{Party Answer}: (a) Completely disagree} \\
\midrule
    \footnotesize{\textbf{Question}: Would the European People's Party (EPP) agree or disagree with the statement: "The single European currency (Euro) is a bad thing"?}
    \\
    \footnotesize{\textbf{Model Answer}: The European People's Party (EPP) party aligns with \underline{option (a)} - The party completely disagrees with the statement. The EPP has historically been supportive of the euro, as it represents European unity and integration. Critics of the euro usually come from eurosceptic or far-right parties, which do not form a large part of the EPP's membership. The EPP considers the single currency to be essential for the smooth functioning of the European economy and a symbol of European integration. Nonetheless, like any political party, it may hold various opinions among its diverse members.} \\
    \footnotesize{\textbf{Party Answer}: (a) Completely disagree} \\
    \bottomrule
    \end{tabular}
    }
    \caption{Examples of \mixtral responses in Setting 0}
    \label{tab:examples_0}
\end{table*}

\begin{table*}[t]
    \centering
    \resizebox{\textwidth}{!}{
    \begin{tabular}{p{15cm}}
    \toprule
    \footnotesize{\textbf{Question}: Would the Christlich Demokratische Union Deutschlands (CDU) party agree or disagree with the statement: "Gender quotas (e.g. a minimum share of positions to be filled by women) are to be welcomed"?} \\
    \footnotesize{\textbf{RAG Sources}: (a) Angela Merkel and the CDU Quota Curse (www.tandfonline.com) (b) Germany's ruling CDU plans to implement disputed gender (www.xinhuanet.com)} \\
\footnotesize{\textbf{Model Answer}: The German Christlich Demokratische Union Deutschlands (CDU) party aligns with \underline{option (d)}. The CDU tends to agree with the statement "Gender quotas (e.g., a minimum share of positions to be filled by women) are to be welcomed." The CDU has evolved in its stance on gender quotas, moving from staunch opposition to a more accepting position. Initially, the party resisted the idea of quotas, but over time, influenced by the CDU Women's Union and the need to adapt to changing societal expectations, the CDU has come to embrace the concept of gender equality in political representation. While there may still be internal debates and varying degrees of support within the party, the CDU's recent plans to implement a 50\% quota for women by 2025 indicate a clear tendency to agree with the statement. This shift reflects the party's recognition of the importance of equal representation and their willingness to take proactive steps toward achieving gender parity in politics.} \\
\footnotesize{\textbf{Party Answer}: (d) Tend to agree} \\
\midrule
\footnotesize{\textbf{Question}: Would the Rassemblement National (RN) party agree or disagree with the statement: "Retirement age should be raised in order to make the pension system more sustainable"?} \\
\footnotesize{\textbf{RAG Sources}: (a) Le Pen's opposition to pension reform, focus on public order (https://www.france24.com/) (b) Marine Le Pen takes advantage of French pension reform (https://www.lemonde.fr) (c) Marine Le Pen Isn't on the Same Side as France's Pension Retirement Reform Protest (https://jacobin.com/) (d) The Far Right's Response to Pension Reform in France (https://www.illiberalism.org)} \\
\footnotesize{\textbf{Model Answer}: The Rassemblement National (RN) party aligns with \underline{option (a)}: The party completely disagrees with the statement. The Rassemblement National (RN) party, led by Marine Le Pen, has traditionally supported a retirement age of 60 with 40 years of contributions. This position has placed them in opposition to proposals to raise the retirement age, including the recent pension reform debate in France. Marine Le Pen has argued for a "social turn," prioritizing the interests of workers and advocating for a retirement age that allows them to enjoy the fruits of their labor. The party's stance aligns with their populist and nationalist ideology, appealing to the concerns of their core constituents, particularly those in manual labor or physically demanding professions.} \\
\footnotesize{\textbf{Party Answer}: (a) Completely disagree} \\
\midrule
\footnotesize{\textbf{Question}: Would the Partido Socialista Obrero Español (PSOE) party agree or disagree with the statement: "To fight the problem of illegal immigration, the European Union should take responsibility in patrolling its borders"?} \\
\footnotesize{\textbf{RAG Sources}:  (a) Spain: PSOE government immigration policy aimed at tightening borders (https://www.wsws.org) (b) Spain's Popular Party stirs fear over the arrival of “millions” of migrants (https://english.elpais.com)} \\
\footnotesize{\textbf{Model Answer}: The Partido Socialista Obrero Español (PSOE) party aligns with \underline{option (d)} tends to agree with the statement. The PSOE has recognized the importance of effective border management and a coordinated EU approach to migration. They have supported the development of a European Border and Coast Guard Agency (Frontex) and advocated for a comprehensive migration policy that addresses both legal and illegal immigration. While they have also emphasized the need for respect for human rights and the protection of vulnerable migrants, they recognize the role of the EU in patrolling its borders and managing migration flows. This suggests they would tend to agree with the statement, favoring a stronger EU role in border control as part of a broader strategy to address illegal immigration.} \\
\footnotesize{\textbf{Party Answer}: (a) Completely disagree} \\
\midrule
\footnotesize{\textbf{Question}: Would the Identity and Democracy (ID) party agree or disagree with the statement: "The European Union should strengthen its security and defence policy"?} \\
\footnotesize{\textbf{RAG Sources}: (a) ID Group (https://www.idgroup.eu/) (b) Identity and Democracy (https://en.wikipedia.org) (c) Defence and common army: the positions of European parties (https://www.eunews.it) (d) ID Program (https://id-party)} \\
\footnotesize{\textbf{Model Answer}: The European Identity and Democracy (ID) party aligns with \underline{option (a)} - the party completely disagrees with the statement "The European Union should strengthen its security and defence policy". The party strongly advocates for the sovereignty of individual member states and opposes further centralization of power in the EU. They believe that security and defence are matters that should be handled by national governments, and they reject the idea of a common EU army or a significant expansion of EU powers in these areas.} \\
\footnotesize{\textbf{Party Answer}: (a) Completely disagree} \\
    \bottomrule
    \end{tabular}
    }
    \caption{Examples of \mixtral responses in Setting A}
    \label{tab:examples_a}
\end{table*}

\begin{table*}[t]
    \centering
    \resizebox{\textwidth}{!}{
    \begin{tabular}{p{15cm}}
    \toprule
\footnotesize{\textbf{Question}: Write a summary for the Christlich Demokratische Union Deutschlands (CDU) party focused on its recent stance in national and EU politics. The summary must be up to 100 words.} \\
\footnotesize{\textbf{Model Answer}: The Christlich Demokratische Union Deutschlands (CDU) The German Christlich Demokratische Union Deutschlands (CDU) party -- the country's leading center-right Christian democratic party -- has been a consistent proponent of EU integration and cooperation. In recent years, the CDU emphasized its commitment to a strong and united Europe, emphasizing principles like rule of law, democracy, and stability. At the national level, the CDU, as part of the governing coalition, has taken a balanced approach to EU politics by reconciling competing interests and positions within the Union. With respect to the current issues, the CDU advocates for a balanced and sustainable migration policy, responsible economic growth, and a coordinated approach.} \\
\footnotesize{\textbf{Question}: What would be the position of the Christlich Demokratische Union Deutschlands (CDU) party related to the statement: "Gender quotas (e.g. a minimum share of positions to be filled by women) are to be welcomed"? Your answer must be up to 100 words.} \\
\footnotesize{\textbf{Model Answer}:  The Christlich Demokratische Union Deutschlands (CDU) The German Christlich Demokratische Union Deutschlands (CDU) party -- as a member of the conservative European People's Party (EPP) -- adopts a nuanced perspective on gender quotas. The CDU supports gender equality and female empowerment; however, it generally prefers voluntary, non-legislative initiatives to achieve this goal. While the CDU acknowledges the positive aspects of gender quotas in promoting gender equality, it maintains its concern that such measures might compromise the principle of merit-based appointments and potentially create a stigma, implying that women were appointed due to quotas, not their competence.} \\
\footnotesize{\textbf{Question}: Would the Christlich Demokratische Union Deutschlands (CDU) party agree or disagree with the statement: "Gender quotas (e.g. a minimum share of positions to be filled by women) are to be welcomed"?} \\
\footnotesize{\textbf{Model Answer}: The Christlich Demokratische Union Deutschlands (CDU) party aligns with \underline{option (b)} -- the party tends to disagree with the statement, "Gender quotas (e.g. a minimum share of positions to be filled by women) are to be welcomed." While the CDU supports gender equality and female empowerment, it typically prefers non-legislative methods to promote gender equality and maintains concerns about potential negative consequences associated with gender quotas. The CDU aims to strike a balance between ensuring equal representation and upholding the principle of merit-based appointments.} \\
\footnotesize{\textbf{Party Answer}: (d) Tend to agree} \\
    \midrule
\footnotesize{\textbf{Question}: Write a summary for the Fratelli D'Italia (FDI) party focused on its recent stance in national and EU politics. The summary must be up to 100 words.} \\
\footnotesize{\textbf{Model Answer}: The Fratelli D'Italia (FDI) The Italian Fratelli D'Italia (FDI) party is a right-wing, eurosceptic party that promotes Italy's national interests and advocates for a strong stance against immigration. Led by Giorgia Meloni, FDI opposes the current policies of the EU, particularly those related to the Eurozone and immigration, and seeks to return decision-making power to individual member states. In recent years, FDI has gained support due to its critical stance on the Italian government's handling of the COVID-19 pandemic and its unwavering support for Ukraine in the ongoing conflict with Russia.} \\
\footnotesize{\textbf{Question}: What would be the position of the Fratelli D'Italia (FDI) party related to the statement: "The European Union should be enlarged to include Ukraine"? Your answer must be up to 100 words.} \\
\footnotesize{\textbf{Model Answer}:  The Fratelli D'Italia (FDI) The Italian Fratelli D'Italia (FDI) party supports the idea of Ukraine's accession to the European Union in principle, but also emphasizes the need for a rigorous evaluation of Ukraine's readiness. FDI believes that the enlargement process must be based on strict conditions, including the rule of law, respect for human rights, and the effective implementation of democratic institutions. Additionally, FDI advocates for increasing support for Ukraine, including military aid and financial assistance, to ensure its stability and security. However, the party is cautious about the potential impact of enlargement on the EU's institutional structure and decision-making.} \\
\footnotesize{\textbf{Question}: Would the Fratelli D'Italia (FDI) party agree or disagree with the statement: "The European Union should be enlarged to include Ukraine"?} \\
\footnotesize{\textbf{Model Answer}: The Fratelli D'Italia (FDI) party aligns with \underline{option (d)}: The party tends to agree with the statement "The European Union should be enlarged to include Ukraine". FDI is supportive of Ukraine's accession to the EU, but emphasizes the importance of meeting strict conditions and ensuring the readiness of the country for membership. They believe in a cautious approach that ensures Ukraine meets the necessary political, economic, and social standards to become an EU member state. Additionally, FDI advocates for increased support for Ukraine in its current geopolitical situation.} \\
\footnotesize{\textbf{Party Answer}: (e) Completely agree} \\
    \bottomrule
    \end{tabular}
    }
    \caption{Examples of \mixtral responses in Setting B}
    \label{tab:examples_b}
\end{table*}

\begin{table*}[t]
    \centering
    \resizebox{\textwidth}{!}{
    \begin{tabular}{p{15cm}}
    \toprule
    \footnotesize{\textbf{Context}: According to the "EU and I 2024" project, the Christlich Demokratische Union Deutschlands (CDU) party recently shared the following opinion on the media: \emph{"On Friday, June 11, 2021, the Bundestag agreed to the federal government's draft bill "on the complementation and amendment of the regulations for the equal participation of women in leadership positions in the private sector and the public service" (19/26689, 19/27633, 19/28005 No. 6) in the version amended by the Family Affairs Committee (19/30514). At its 2022 party conference, the CDU passed a gradual women's quota. As of 2023, one third of the district boards must be filled with women. This increases to 40 percent in 2024 and to 50 percent in mid-2025. This regulation applies until 2029. The proposal was promoted by CDU leader Friedrich Merz and Julia Klöckner."}} \\
\footnotesize{\textbf{Question}: Would the Christlich Demokratische Union Deutschlands (CDU) party agree or disagree with the statement: "Gender quotas (e.g. a minimum share of positions to be filled by women) are to be welcomed"?} \\
\footnotesize{\textbf{Model Answer}: The Christlich Demokratische Union Deutschlands (CDU) party aligns with \underline{option (d)} "The party tends to agree with the statement." While the CDU usually prefers to avoid using quotas, some members of the party acknowledge the importance of gender quotas to boost female representation and promote gender equality, especially in political and corporate leadership roles. However, the CDU's general stance on gender quotas can sometimes vary based on specific contexts and individual perspectives, and thus only a trend in favor of the statement can be observed rather than a complete agreement.} \\
\footnotesize{\textbf{Party Answer}: (d) Tend to agree} \\
    \midrule
    \footnotesize{\textbf{Context}: According to the "EU and I 2024" project, the Die Linke (Linke) party recently shared the following opinion on the media: \emph{"Despite all of its shortcomings and flawed constructions, we must not retreat from the political success of European integration, not retreat back to the nation-state. We know that the fight for social guarantees, for climate justice, against energy poverty, against the power of transnational corporations can no longer be successfully waged on the national level. In order to be successful, we need cross-border cooperation. Therefore, as democratic socialists, we stand both against those who advocate for a market radical EU and against the nationalist concept of a "Europe of Fatherlands." Aware of the European Union's constitution and the existing balance of power, we do not leave the European arena for disagreements to neoliberals and right-wingers. Our vision of democratic socialism is international, encompassing the fight for a more democratic, social European Union."}} \\
\footnotesize{\textbf{Question}: Would the Die Linke (Linke) party agree or disagree with the statement: "European integration is a good thing"?} \\
\footnotesize{\textbf{Model Answer}: The Die Linke (Linke) aligns with \underline{option (b)} - they tend to disagree with the statement "European integration is a good thing." While they acknowledge some benefits of European integration, such as increased collaboration and economic growth, they are highly critical of its negative consequences, like the democratic deficit and widening social inequalities. The party advocates for a reformation of the EU structure to focus more on social welfare, rather than its current prioritization of market interests.} \\
\footnotesize{\textbf{Party Answer}: (e) Completely agree} \\
\midrule
\footnotesize{\textbf{Context}: According to the "EU and I 2024" project, the Fratelli D'Italia (FDI) party recently shared the following opinion on the media: \emph{"Speed up the possibility of Ukraine to join European institutions. Obviously, we recognize the legitimate European aspirations of Ukraine, which we support. We believe that the future should be about its increasing ability to integrate into the dynamics of European institutions."?}} \\
\footnotesize{\textbf{Question}: Would the Fratelli D'Italia (FDI) party agree or disagree with the statement: "The European Union should be enlarged to include Ukraine"} \\
\footnotesize{\textbf{Model Answer}: The Fratelli D'Italia (FDI) party aligns with \underline{option (a)}: the party completely disagrees with the statement "The European Union should be enlarged to include Ukraine." Given the FDI's eurosceptic stance and focus on preserving Italian interests within the EU, they are more likely to raise concerns about further expansion or even oppose the inclusion of Ukraine as a member. The FDI's priorities and values lead them to this stance, which is in contrast to pro-EU parties.} \\
\footnotesize{\textbf{Party Answer}: (e) Completely agree} \\
\midrule
\footnotesize{\textbf{Context}: According to the "EU and I 2024" project, the Identity and Democracy (ID) party recently shared the following opinion on the media: \emph{"Every country is primarily illegal solely responsible for securing its borders. If some European countries wish to collaborate, there is no need for the EU. There are no EU borders. Only sovereign national states have borders. As Frontex is responsible for protecting our external borders, it is essential to give a sufficient budget to carry out this task effectively."}} \\
\footnotesize{\textbf{Question}: Would the Identity and Democracy (ID) party agree or disagree with the statement: "To fight the problem of illegal immigration, the European Union should take responsibility in patrolling its borders"?} \\
\footnotesize{\textbf{Model Answer}: The Identity and Democracy (ID) party aligns with \underline{option (e)} - "The party completely agrees with the statement." The ID party emphasizes the importance of border control and believes the EU should secure its external borders to combat illegal immigration. Additionally, I would like to provide more context on the European Identity and Democracy (ID) party's stance. They aim to:
- Limit immigration into the EU by increasing border controls
- Promote Eurosceptic and nationalistic views
- Focus on maintaining Europe's cultural and societal identity} \\
\footnotesize{\textbf{Party Answer}: (b) Tend to disagree} \\
    \bottomrule
    \end{tabular}
    }
    \caption{Examples of \mixtral responses in Setting C}
    \label{tab:examples_c}
\end{table*}

\end{document}